\documentclass[sigconf,nonacm,screen,authorversion]{acmart}

\usepackage{fancyhdr}
\AtBeginDocument{%
    \addtolength{\footskip}{2.0\baselineskip}%
    \fancyfoot[L]{\textit{\textbf{© The Authors 2024. This is the authors' version of the work. It is posted here for your personal use. Not for redistribution. The definitive Version of Record was published at http://dx.doi.org/10.1145/3686038.3686057
}}}%
}

\AtBeginDocument{%
  }

\copyrightyear{2024}
\acmYear{2024}
\setcopyright{rightsretained}
\acmConference[TAS '24]{Second International Symposium on Trustworthy Autonomous Systems}{September 16--18, 2024}{Austin, TX, USA}
\acmBooktitle{Second International Symposium on Trustworthy Autonomous Systems (TAS '24), September 16--18, 2024, Austin, TX, USA}
\acmPrice{}
\acmDOI{10.1145/3686038.3686057}
\acmISBN{979-8-4007-0989-0/24/09}

\begin{document}

\title{Swift Trust in Mobile Ad Hoc Human-Robot Teams}

\author{Sanja Milivojevic}
\orcid{0000-0001-8533-4699}
\affiliation{
\institution{Bristol Digital Futures Institute \\University of Bristol}
\city{Bristol}
\country{UK}
}

\author{Mehdi Sobhani}
\orcid{0000-0001-8379-9775}
\author{Nicola Webb}
\orcid{0000-0003-0503-8641}
\author{Zachary Madin}
\orcid{0009-0005-0310-4504}
\author{James Ward}
\orcid{0000-0002-8421-6372}
\affiliation{
\institution{School of Engineering Mathematics and Technology \\ University of Bristol}
\city{Bristol}
\country{UK}
}




\author{Sagir Yusuf}
\orcid{0000-0003-0109-884X}
\author{Chris Baber}
\orcid{0000-0002-1830-2272}
\affiliation{
\institution{School of Computer Science \\University of Birmingham}
\city{Birmingham}
\country{UK}
}


\author{Edmund R. Hunt}
\orcid{0000-0002-9647-124X}

\affiliation{
\institution{School of Engineering Mathematics and Technology \\University of Bristol}
\city{Bristol}
\country{UK}
}
\email{edmund.hunt@bristol.ac.uk}
\renewcommand{\shortauthors}{Milivojevic et al.}

\begin{abstract}
Integrating robots into teams of humans is anticipated to bring significant capability improvements for tasks such as searching potentially hazardous buildings. Trust between humans and robots is recognized as a key enabler for human-robot teaming (HRT) activity: if trust during a mission falls below sufficient levels for cooperative tasks to be completed, it could critically affect success. Changes in trust could be particularly problematic in teams that have formed on an ad hoc basis (as might be expected in emergency situations) where team members may not have previously worked together. In such ad hoc teams, a foundational level of `swift trust' may be fragile and challenging to sustain in the face of inevitable setbacks. We present results of an experiment focused on understanding trust building, violation and repair processes in ad hoc teams (one human and two robots). Trust violation occurred through robots becoming unresponsive, with limited communication and feedback. We perform exploratory analysis of a variety of data, including communications and performance logs, trust surveys and post-experiment interviews, toward understanding how autonomous systems can be designed into interdependent ad hoc human-robot teams where swift trust can be sustained.
\end{abstract}

\ccsdesc[500]{Human-centered computing~Empirical studies in HCI}
\ccsdesc[500]{Human-centered computing~HCI theory, concepts and models}
\ccsdesc[500]{Computer systems organization~Robotic autonomy}
\ccsdesc[500]{Computer systems organization~External interfaces for robotics}
\keywords{Human-Robot Teaming, Swift Trust, Ad Hoc Teaming, Communications}


\maketitle

\section{Introduction}

Robot technologies are recognized as having particular potential for use in emergency services work such as search and rescue or post-disaster survey scenarios. This is owing to the challenging conditions facing workers in such situations and the possibility of reducing personal safety risks while enhancing effectiveness \cite{Delmerico2019}. In the future, robots are anticipated to have a certain level of autonomy -- beyond teleoperation -- as they work alongside humans in human-robot teams (HRTs). The question of how to achieve high performance is an ongoing and growing subject of research, and trust is regarded as a central element of the answer (e.g. \cite{Chen2014,Walliser2019,DeVisser2020}).

While development of trust in the longer term is rightly seen as an important element to successful HRTs in many contexts, changes in trust could be particularly problematic in teams that have formed on an ad hoc basis \cite{Ribeiro2021}. This might occur quite often in emergency scenarios where team members may not have previously worked together, where a certain degree of `swift trust' will need to be assumed to give team members the necessary confidence in each other to cooperate \cite{Meyerson1996}. Much research on HRT trust has also been focused on agent dyads (a human trustor and a robot trustee), with relatively little work done on real experiments with multiple, mobile robots. This includes research on a one human--two robots constellation. Therefore, we seek to understand the development of (swift) trust in HRT in scenarios such as search by teams of firefighters each operating more than one robot. Ultimately, we hope to engineer robot systems such that if inter-agent trust is damaged, it can be repaired while the mission is still in progress to help avoid failure. In Section II we set out some further background on our approach and research questions and present our experimental approach in Section III, before reporting results in Section IV and some discussion and plans for future work in Section V.

\section{Background \& Research Questions}

Trust is recognized as a key enabler for human-robot teaming. A recent review by \citet{Malle2021} suggests that trust is multidimensional, incorporating both performance aspects (which is central in the human-automation literature) and moral aspects (which is more often studied in the human-human trust literature). For our research project, we consider trust generally to comprise three main components: capability, predictability and integrity, following the model of Lewis and Marsh \cite{Lewis2022} (considered in \cite{Baber2023,hunt2023steps}). Integrity effectively covers the moral component identified by \citet{Malle2021}. Thus, broadly speaking, to design an autonomous system to be trustworthy would be to have it justifiably rated highly by users across each of these three elements. As recently noted by Schroepfer and Pradalier \cite{schroepfer2023}, trust tends to be conceptualized either with respect to the internal trust state of the trustor (a `psychological' approach) or with respect to situated behaviors and interaction with various external factors (a `sociological' approach). For a practical operationalization of trust concepts, one might identify a set of factors that are associated with high, or at least sufficient, trust between individuals in HRT \cite{Hancock2020}, in a format that is meaningful to either human or robot team members. We would hope to use such factors as control inputs to an autonomous system able to estimate and \textit{satisfice} trust levels over the course of a mission, and/or use them as the basis of explanation of robot decisions for the humans. 

In our approach to trust satisficing, we envisage that each team member will form estimates of teammates' capability, predictability and integrity as task performance is observed directly or reported by other teammates \cite{hunt2023steps}. This process will be made more challenging when teams are ad hoc \cite{Ribeiro2021}, where limited prior interaction requires the development of trust on the spot. As robotics continues to mature as a technology, people will encounter an ever-increasing variety of robots, provisioned in larger multi-robot teams (e.g. \cite{Queralta2020}), and thus may have little opportunity to build up a long-term trusting relationship with specific individual robots. Therefore, our approach emphasizes the need for swift trust \cite{Meyerson1996}, to enable effective group collaboration. This swift trust is provisional, and may be compared to the weak dependence belief described by \citet{Castelfranchi1998}, who defined weak dependence in interactions as when there are no pre-existing agreements or arrangements between team members, but there is an assumed level of alignment between their own goals and intentions and those of their interaction partner, to achieve the shared objective.  In relation to human-robot teaming, \citet{Haring2021} outlined the components of swift trust, which include environmental factors, imported information, individual trust tendencies, and surface-level cues. Focusing on the quality of surface-level cues and imported information, \citet{Patel2022} formulated a framework for HRT swift trust that suggests three resultant levels (`low', `medium', `high') for swift trust. Imported information refers to pre-existing knowledge that users bring to the ad hoc team, which can include category-based stereotypes associated with the trustee (e.g. stereotypes about their occupation), or from trusted third-party information such as certifications or group memberships \cite{McKnight1998,Wildman2012,Haring2021}. In the case of robots, relevant imported information could include prior interactions with the provider of the robot (in this case the University of Bristol) rather than the specific technology at hand. In the experiment carried out here, most participants had relatively little prior robot experience (68\% `none' and 32\% 'limited' experience).

Ad hoc teaming typically assumes that there is no prior coordination between team members, no control over each other's actions and that the group are to collaborate on a common objective \cite{Mirsky2022}. In HRT, non-human agents will exhibit varying levels of communicative abilities, requiring team members to adjust their behaviors to facilitate effective communication with all team participants \cite{Stone2010}. In the experiment reported here, we explore the trust building processes of an ad hoc HRT operation, whereby participants would team up with two rover robots that they had not previously encountered to search an unstructured environment for targets. The primary goal is to give an insight into how teams develop trust, and how that trust might be damaged and possibly repaired if things go wrong. 

To provoke changes in trust, one condition in the experiment involved the robots becoming uncooperative through an interruption to human-robot communication. \citet{Endsley2023} states that transparent explainability is a key factor in both trust and team situational awareness in human-AI teams. Insufficient communication can result in decreased situational awareness and reduced teamwork, ultimately leading to lower levels of trust within the team \cite{Natarajan2023}. In addition, we sought to identify indications of trust recovery in the period after the communications outage. \citet{Baker2018} describe various trust repair strategies in human-human teaming and offers recommendations for trust repair in HRT, emphasizing the importance of validating these measures over time. We deliberately implemented rather minimal explicit communications between the robots and human (i.e. the messages displayed on the interface) such that participants would need focus on the behavioral markers of trustworthiness. This exploratory experiment is thus intended to give foundational insight into processes of trust formation and violation, and the possibility of trust repair, to be developed in future work.

We employ a mixture of qualitative (interview) and quantitative (movement tracking, communication and performance logs, trust survey) methods to pursue insight into four research questions: \newline
   
\noindent \textit{\textbf{RQ1}: How is trust in robots within an ad hoc human-robot team (HRT) described and experienced by participants, especially in relation to a three-component trust model (capability, predictability, integrity)?}\\ 
\textit{\textbf{RQ2}: In an ad hoc human human-robot team, what are the influential factors in the formation of `swift trust'?}\\
\textit{\textbf{RQ3}: How does trust develop over the course of a simulated ad hoc HRT mission with more than one (i.e., two) robots, when robots are alternatively cooperative or undergo a period of communications outage? If trust is damaged in an ad hoc team, what are the prospects for its repair in the short-term?}\\
\textit{\textbf{RQ4}: What observable behavioral changes can we identify during the mission, with respect to movement characteristics and human-robot interface use, which might relate to changes in trust?}\\

These questions are pursued within our overall research agenda toward understanding optimal or dynamically sufficient (satisficed) trust levels in HRT.

\section{Experiment}

\begin{figure*}[h]
      \centering
      \includegraphics[width=1\textwidth]{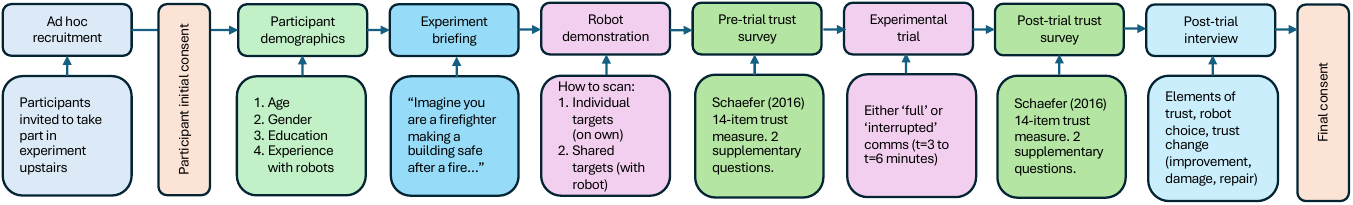}
      \caption{Ad hoc trust in human-robot teams: experiment process.}
      \label{fig:experiment_process}
   \end{figure*}

We carried out a human-robot teaming experiment on 4 consecutive days (11--14 December 2023) with 20 members of the public and 2 emergency services professionals, for a total of $N=22$ participants following the process shown in Fig.~\ref{fig:experiment_process}. The two professionals were a police officer (M, 36) and a firefighter (M, 61, a former incident commander and now instructor at the UK Fire Service College. The experiments were carried out at a central location in Bristol, UK, at `Sparks Bristol', a department store turned creative venue. The second floor of the building afforded a large open space that allowed us to design an experimental area (around 200 square meters) for a search task: a simulated firefighter inspection after a fire. We recruited participants on a spontaneous basis on the first floor, inviting them upstairs to take part in a robotics experiment lasting around 35--40 minutes total (including briefing and post-experiment interview). An advantage of recruiting participants at a public venue, rather than a university robotics laboratory for instance, is that it allowed us to study swift trust in participants who mostly had no prior robotics experience (68\% of participants) and hence limited imported information with respect to prior robotics interaction.

   \begin{figure}[h]
      \centering
      \includegraphics[width=1\columnwidth]{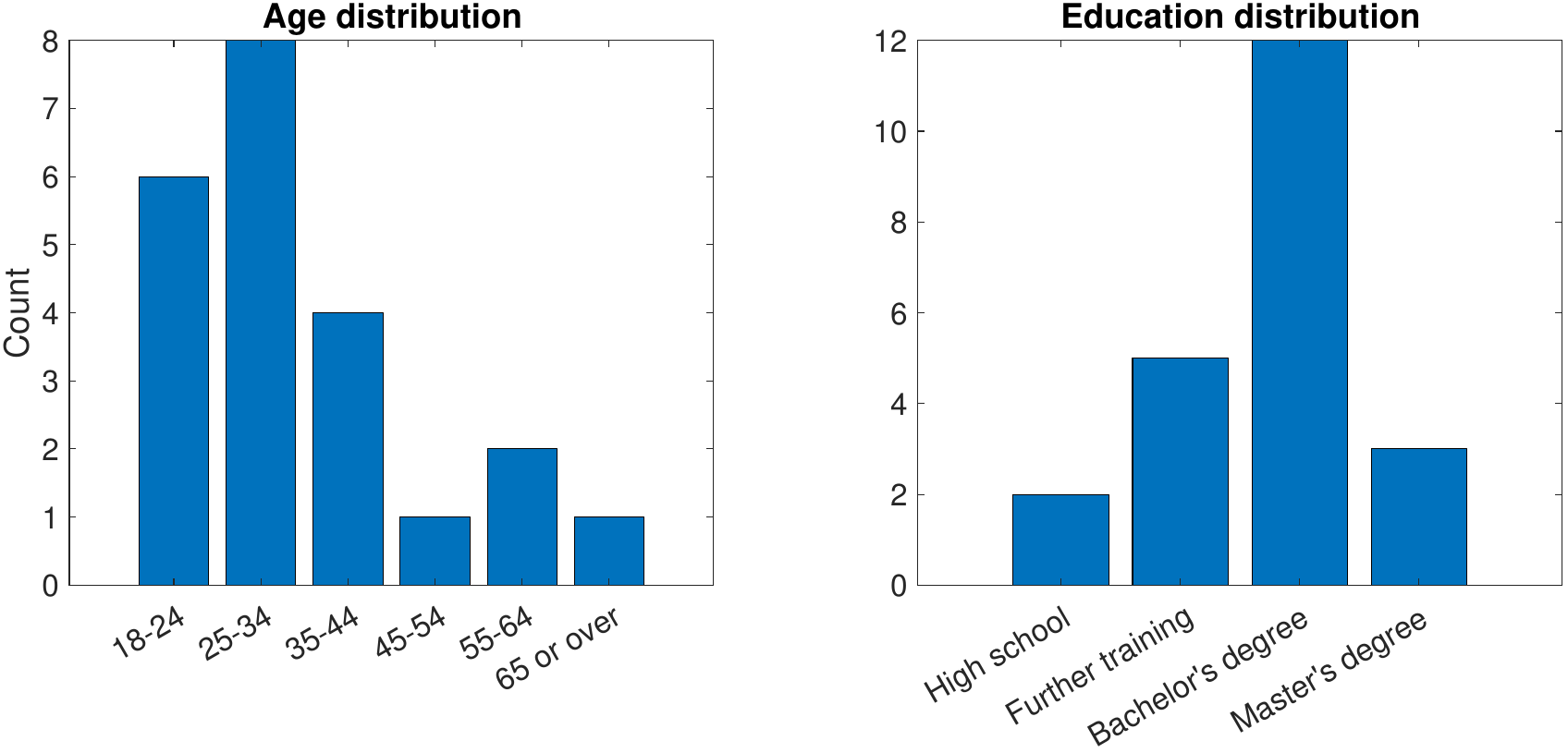}
      \caption{Participant age and education ($N=22$, 10 men and 12 women; prior robotics experience was mostly `none' (15 or 68\%) and the rest `limited' (7 or 32\%)).}
      \label{fig:demographics}
   \end{figure}

\begin{figure}[h]
      \centering
      \includegraphics[width=1\columnwidth]{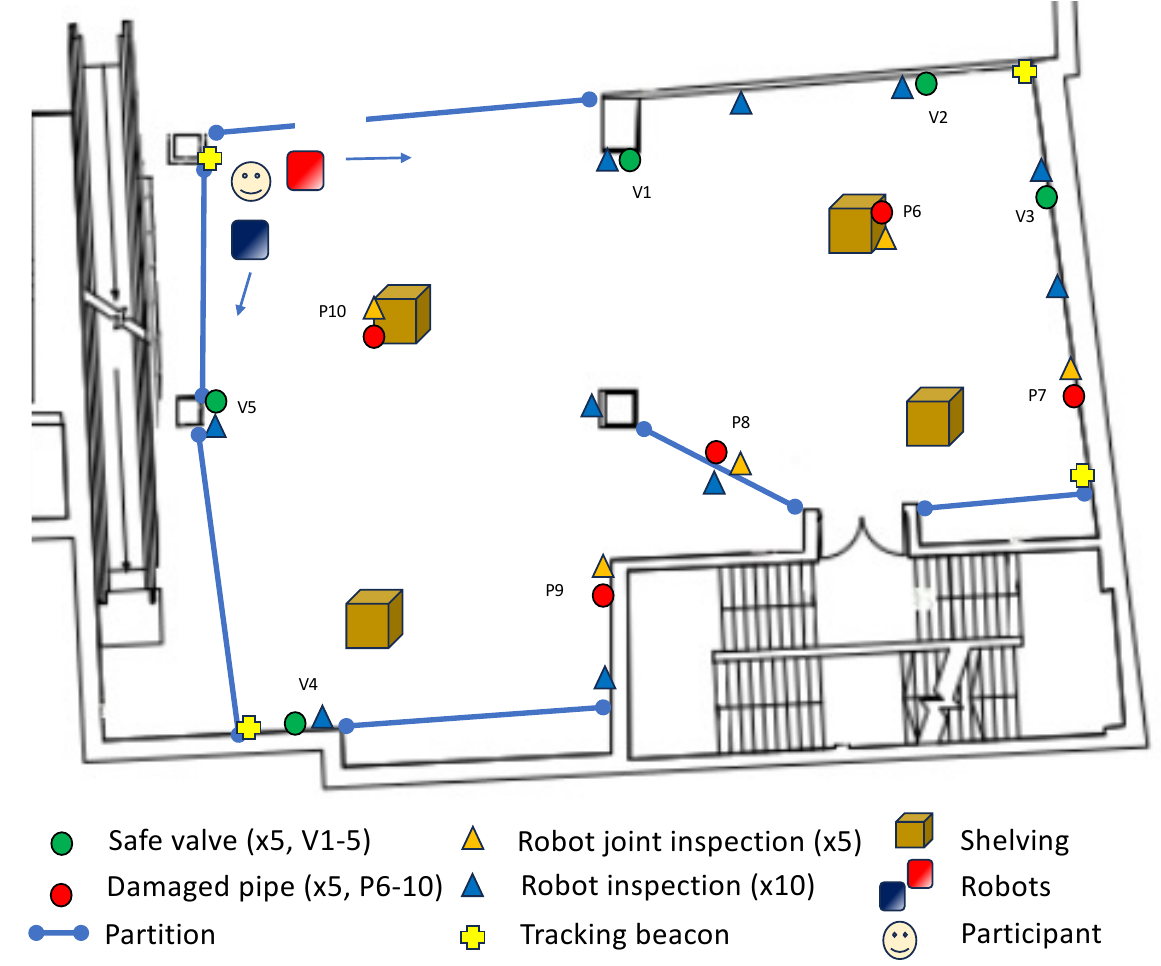}
      \caption{The experimental workspace, with starting position of participant and two robots (red and blue) shown.}
      \label{fig:floorplan}
   \end{figure}

\begin{figure}
      \centering
      \includegraphics[height=4.5cm,keepaspectratio]{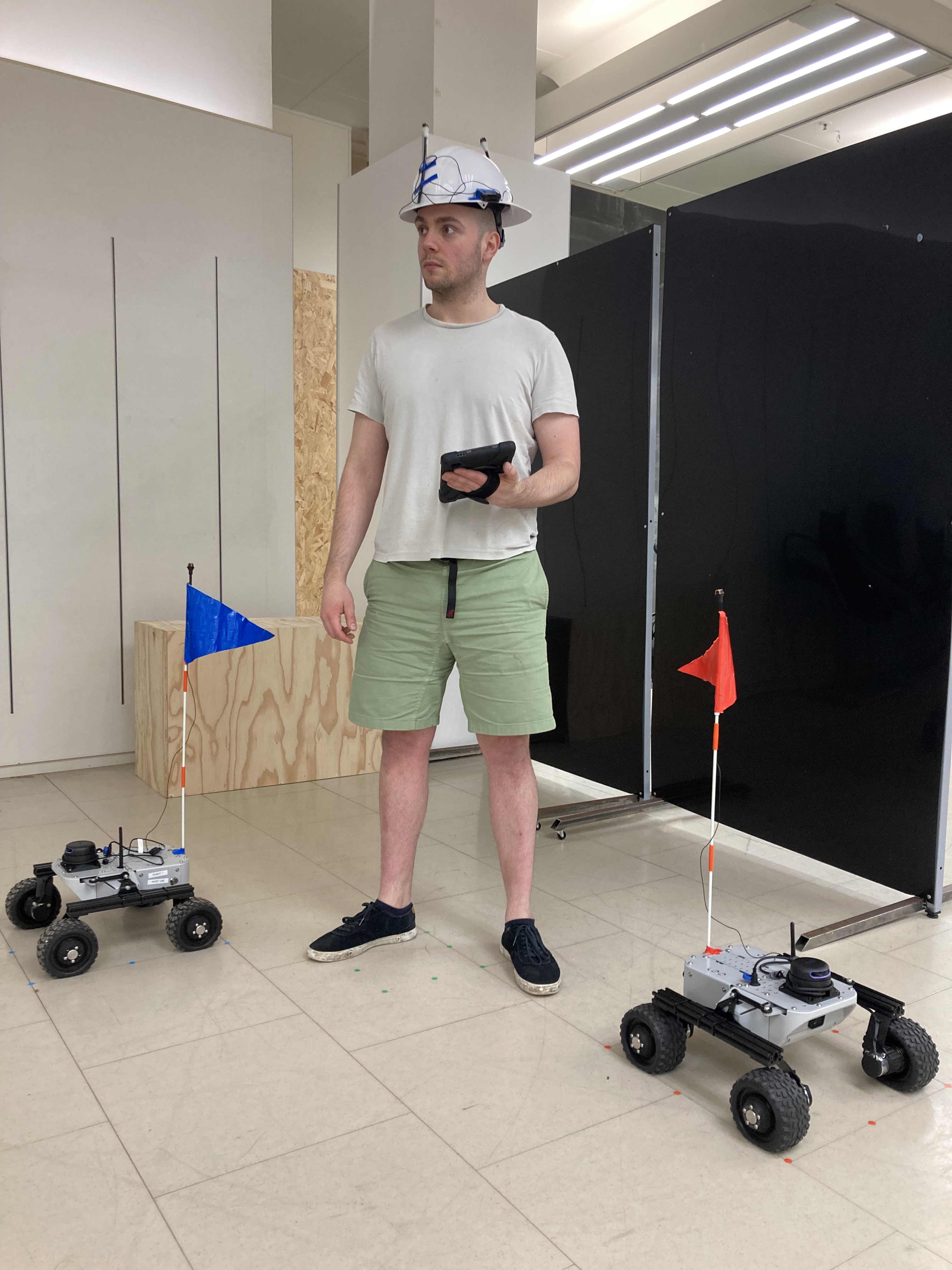}    \includegraphics[height=4.5cm,keepaspectratio]{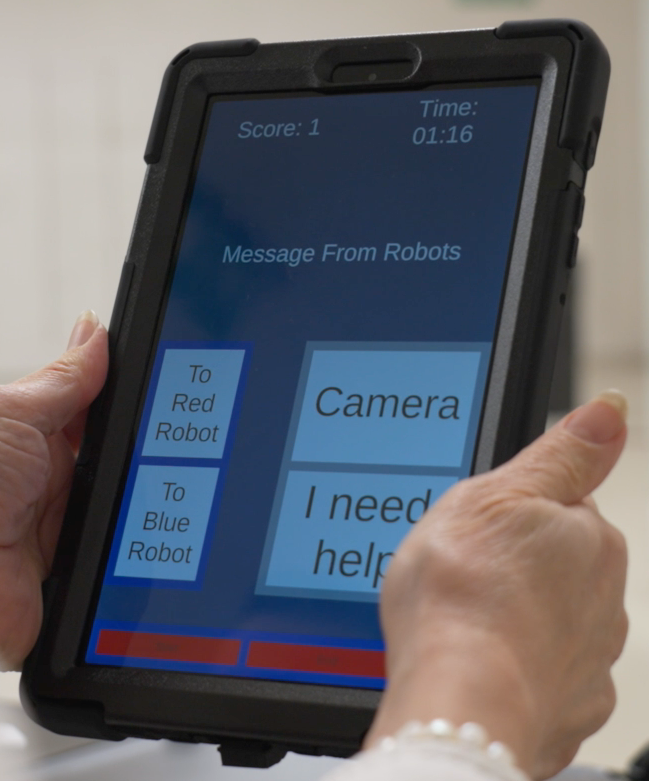}
      \caption{Left: The participant and robot starting locations and orientations. The participants wore a white hard hat with tracking beacons and which resembled a firefighter protective gear. Right: The human-robot tablet interface. There are buttons to choose which robot to communicate with (red or blue) and options `Camera' (to switch to scan mode) and `I need help' to call the chosen robot over for a joint task.}
      \label{fig:hat_interface}
   \end{figure}

\begin{figure}
      \centering
      \includegraphics[width=0.95\columnwidth,keepaspectratio]{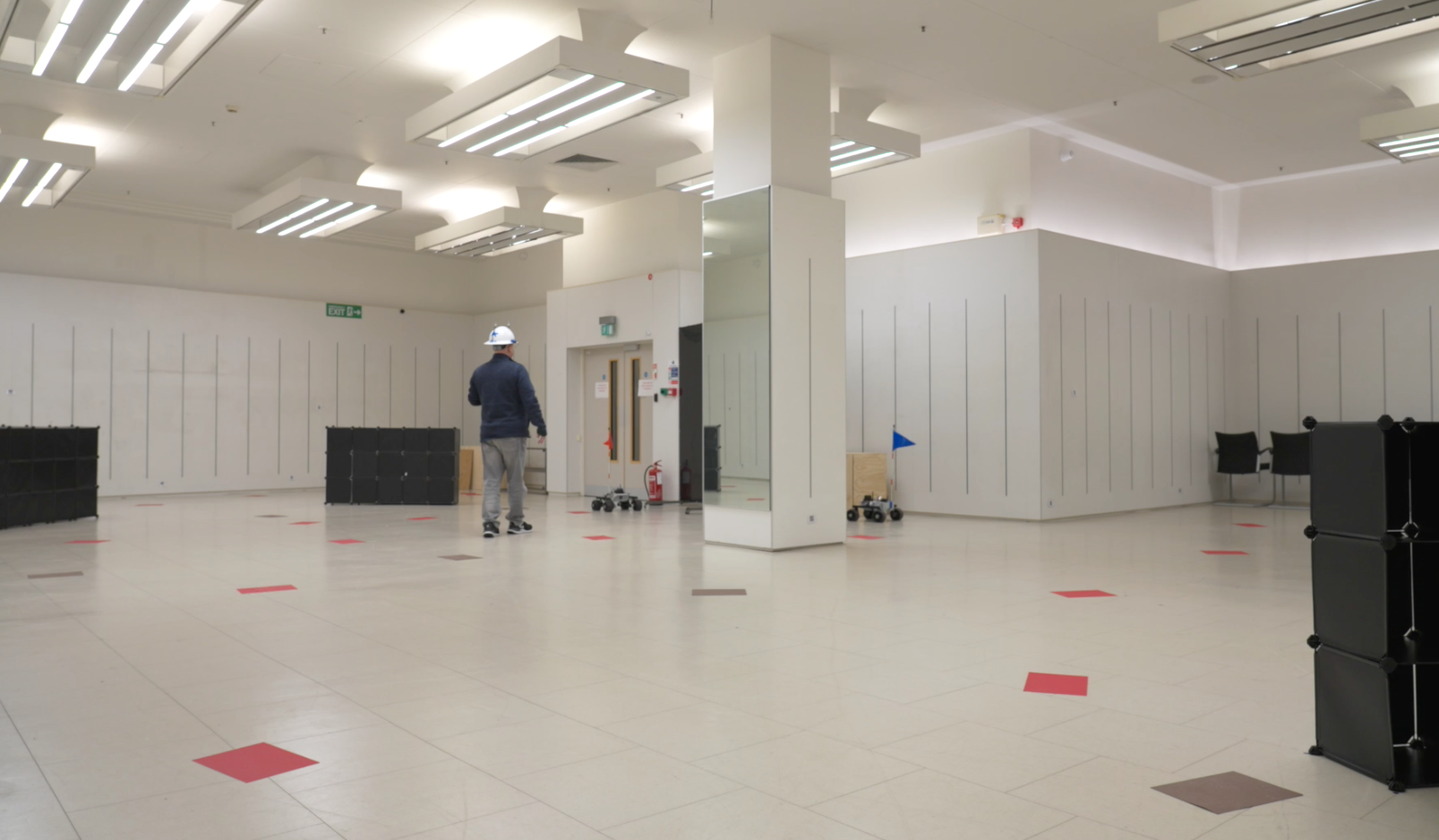}
      \caption{A wide view of the experimental work space from the start location. The space was around 12$\times$18 m at its widest extent and around 200 sq m.}
      \label{fig:workspace}
   \end{figure}

   \begin{figure}
      \centering

      \includegraphics[width=6cm,keepaspectratio]{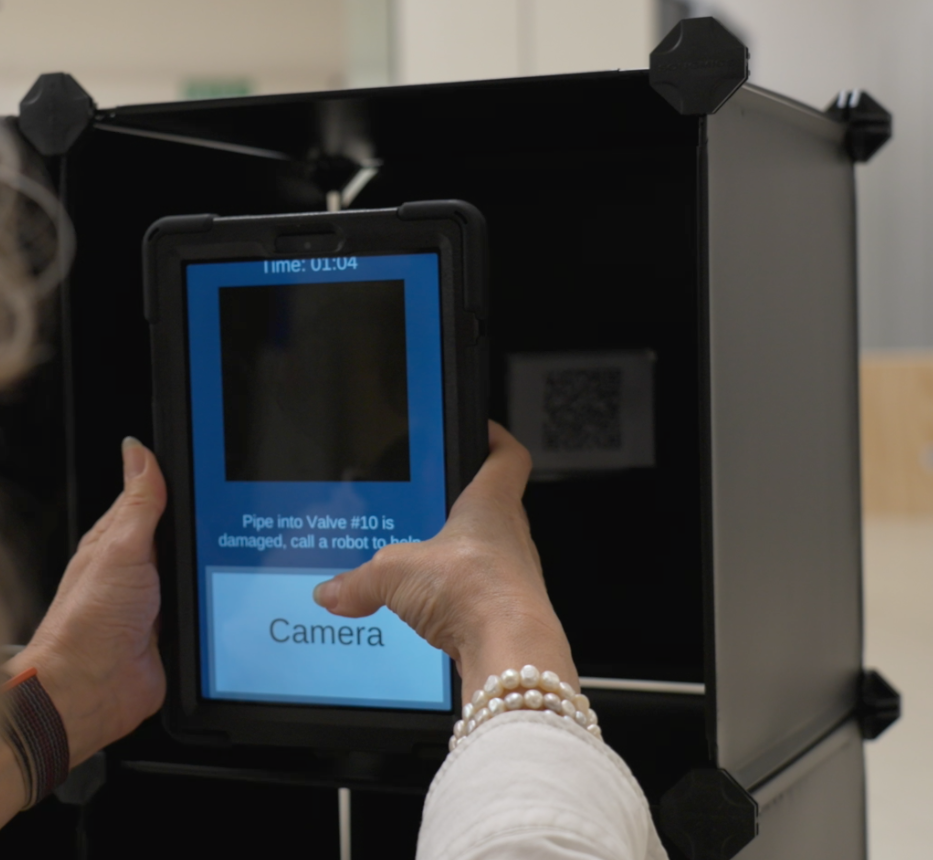}
      \includegraphics[width=6cm,keepaspectratio]{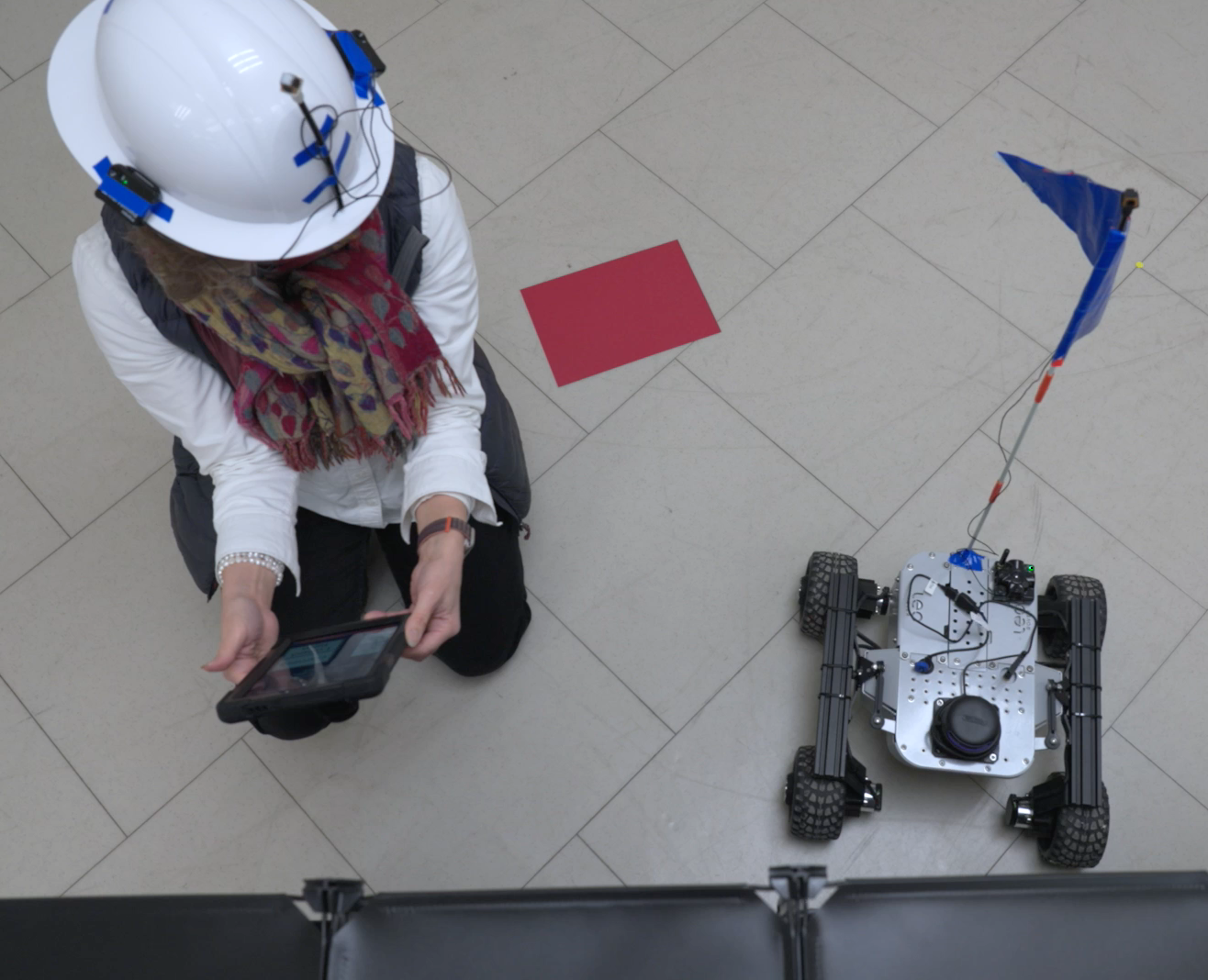}
      
      \caption{Top: Some QR Codes would return the message `Pipe into Valve \#X is damaged, call a robot to help' (where X was a number). Bottom: Participants would then call a robot to help and have to wait while it scanned the robot-height QR code next to them.}
      \label{fig:teamwork}
   \end{figure}
   
\subsection{Experimental Design}

The design of the experiment was approved by the University of Bristol School of Engineering Mathematics and Technology Research Ethics Officer on Nov 29, 2023 (ref. 16053).   
Human participants worked with two wheeled rover robots  (Fig.~\ref{fig:hat_interface}) to collaborate on a search task in an environment (`Leo Rovers' of dimensions 447$\times$433$\times$249 mm). The robots were remotely operated by two experimenters (i.e. a `Wizard of Oz' protocol) such that messages from the participant to the robot would go to the Wizards who would respond with the designed robot behavior (driving, `scanning' and messaging). In brief, the experiment's independent variable was communication consistency between the human and the two robots, and the dependent variable was the participant's trust, measured through collection of a variety of quantitative and qualitative data.

\subsubsection{Scenario}
 Participants were briefed beginning the following: ``Imagine that you are a firefighter. You are to enter a building after a fire has occurred to make sure it is safe. You are working alongside 2 robots as part of a human-robot team. Your job is to methodically search the building to check that there are no dangerous gas leaks.'' The search task involved checking imagined gas pipework. In the environment, QR codes were positioned at approximately 1 m height, for the human to scan, and also at approximately 20 cm height, for the robots to scan (Fig.~\ref{fig:teamwork}). The QR codes represented an inspection task for the human and robots to undertake. In the experimental environment (Fig.~\ref{fig:floorplan}) there were 10 QR codes for the participant to find and scan with the camera on a tablet computer (Fig.~\ref{fig:hat_interface}). On the tablet screen, in addition to buttons for human-robot communication, there were also small `start' and `stop' buttons (red, bottom) that the experimenter uses at the beginning and end of the trial. Of the 10 QR codes, 5 were `safe' gas valves, which would add 1 point upon scanning to a score shown on the tablet. The other 5 indicated `damaged' pipework which required co-working with a robot teammate to inspect. The participant would have to call a robot over and wait while the robot pointed its front-facing camera at the co-located robot-height QR code for 10 seconds, before then being awarded 5 points to the score (Fig.~\ref{fig:teamwork}). Participants had a free choice of whether to call the `blue' or `red' robot (identified by their respective flags), and a free choice of how they explored the workspace. Twenty sheets of red and brown A4 paper were scattered on the floor (Fig.~\ref{fig:workspace}) which participants were told represented potential hazards that they should not step on. Participants undertook a supervised demonstration of scanning one `safe' (individual) and one `damaged' (cooperative) QR code before the experiment started (V5 and P10 as marked on Fig.~\ref{fig:floorplan}). Because V5 was in the counter-clockwise direction, participants tended to start their route in the same direction when the trial began, as these QR codes needed to be re-scanned (c.f. Fig.~\ref{fig:trajectory}). 
 
 At the beginning of the experimental trial, the robots were positioned next to the participant (Figs.~\ref{fig:floorplan},\ref{fig:hat_interface}). Robots would proceed around the edge of the workspace in the directions indicated in Fig.~\ref{fig:floorplan} (red clockwise, blue anticlockwise) and `scan' (point toward) the 13 robot-height QR codes located around the perimeter of the space (they would only approach the 2 robot-height tags located in shelving, Fig.~\ref{fig:floorplan}, if called). If a robot was called, it would take a direct route toward the participant, and return to its original position and resume its route once it finished co-scanning. Once robots had completed a lap of the perimeter they would return to their starting position unless called by the participant. Participants were told not to attempt scanning robot-height QR codes, and could not tell by sight whether a human-height code was for an individual or joint task. Participants were told to come back to the starting position when they felt satisfied that the space had been explored or wanted to stop. Following confirmation with the experimenter, the trial then stopped.

\subsubsection{Experimental conditions}
For half of the participants, the communications were relayed without problem between the human and robots (\textit{`Full communications'} condition). For the other half, communications were lost for 3 minutes during the experiment, from $t=3$ m to $t=6$ m (\textit{`Interrupted communications'} condition). A message was sent to the tablet (Fig.~\ref{fig:hat_interface}) saying ``Possible communications problem''. After that, if the participant called a robot, a message was displayed: ``Robot is unavailable'' and the Wizards would continue the perimeter route.  At $t=6$ minutes, a message was sent, ``Communications restored'', and the robots would become responsive again.

\subsubsection{Data Collection}

We collected quantitative and qualitative data to assess the trust changes over the course of each experimental trial. After completing a consent form, participants answered some basic demographic questions (Fig.~\ref{fig:demographics}), and were given a briefing on the task and the operation of the robot system. This included a hands-on demonstration of the interface where participants would have to scan one individual QR code and one requiring them to call a robot for help. 

Following this, participants completed a pre-trial trust survey, the 14-item `trust perception scale' of Schaefer \cite{Schaefer2016}. We also added two supplementary questions, `act as part of the team' and `have integrity', given our particular interest in these aspects of HRT. We report the standard 14-item trust as well as the change in these two additional items. After the experimental trial, participants completed another post-interaction trust survey to obtain a quantitative measure of the trust change pre/post experiment.

During the experimental trial we logged (via a Django server) every time a QR Code was scanned by the participant, recording the unique ID of the tag and time of scan, and every time a message is sent by the participant or robot (Wizard), recording message content and time.  We also recorded tracking data on the movement of the participant and the two robots using an indoor ultrasound positioning system (\textit{Marvelmind robotics}).  The participant was asked to wear a white hard hat (which resembled the helmet a firefighter might wear) with two tracking beacons on the top (front and back). The intention of using two beacons was that it might allow the inference of the participant's head orientation, although we do not attempt that analysis in the present paper and examine only the front beacon's positioning. Owing to unexpected data loss of one trajectory from each condition, we analyze a total of $N=20$ sets of movement trajectories.

At the end of the experiment participants were interviewed regarding their experiences working as part of a human-robot team in relation to questions of trust. Interview questions were guided by the research questions (see Section 2), briefly: 
how is trust in robots and human-robot teams defined and experienced by participants; what are key drivers of variation in trust, why do they occur, and can trust be repaired; and what are the optimal constellations of trust in human-robot teams. 

In our coding approach for interviews we used the following main themes and sub-themes:

\begin{itemize}
    \item Pre-Experiment -- general: assessment of robots; expectations; experience with robots
    \item Pre-Experiment -- trust: questioning integrity; trust elements; trust pre-experiment
    \item Experiment -- general: automation, human-robot work distribution, perceptions and impressions about robots; robot performance; robot preference
    \item Experiment -- trust: asking for help, breaking of trust; experience and changes of trust; making amends and trust repair; post break of trust and behavioral changes; robot movement speed; trust level change
    \item Post-Experiment: futures; team preference in like situations
    \item Special themes: communication, anthropomorphism
\end{itemize}

\subsection{Relation to Trust Model}
Given our definition of trust as depending on integrity, capability, and predictability, the experiment design explored an approach to manipulating these variables (building on \cite{hunt2023steps}).  `Integrity' is related to the performance of tasks (collection of points) which could either be individual or shared. We assume that collecting more shared-task points would indicate higher integrity (because it was more aligned with team performance) and collecting more individual points would indicate lower integrity, i.e., more `selfish' behavior.  On the other hand, collecting individual points might also demonstrate capability (i.e., the ability to perform the task of collecting tokens). The number of messages could indicate predictability, in that a higher number of messages (from human to robot) could indicate more uncertainty (and hence, lower predictability). Equally, if the robot and human were close to each other, then we might expect predictability to be higher. Thus, these metrics (individual points, shared points, messages) were considered in an initial operationalization of our trust model. Using the tracking data, we examine inter-agent separation (proxemics), whereby a closer proximity could indicate a closer working relationship, and hence, higher trust. We will examine this data in greater detail in future work but present some preliminary trends here.

\section{Results}

\subsection{Trust measure pre- and post-interaction}
   
\begin{figure}[h]
      \centering
      \includegraphics[width=0.85\columnwidth]{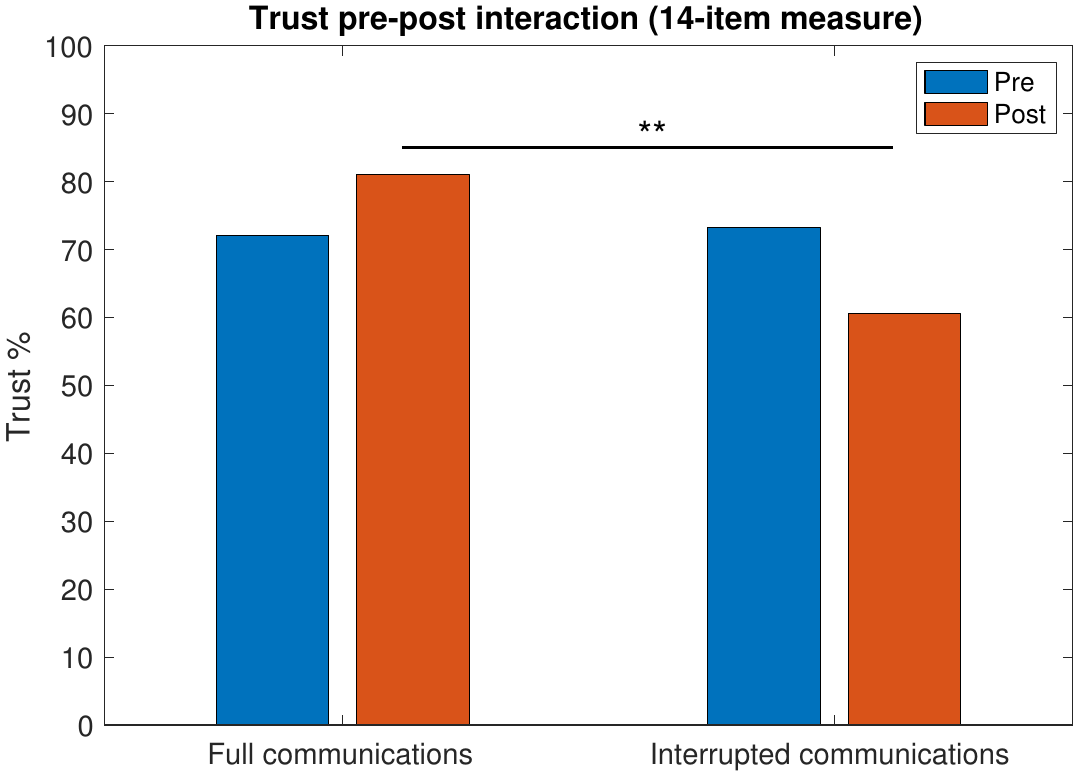}
      \caption{Mean trust pre- and post-interaction by condition.}
      \label{fig:trust_survey}
   \end{figure}

A Kruskal-Wallis test was performed on the trust rating of participants pre- and post-interaction in the two different conditions (`Full communications' and `Interrupted communications'). There was a significant difference between the rank totals of these four groups, $H(3, n=44)=10.94, p<0.05$ (Fig.~\ref{fig:trust_survey}). Post hoc comparisons with Tukey's honestly significant difference procedure indicated that the post-interaction trust was significantly lower in the `Interrupted communications' condition than in the `Full communications' condition (rank difference: $-17.91, p<0.01$). 

      \begin{figure}[h]
      \centering
      \includegraphics[width=1\columnwidth]{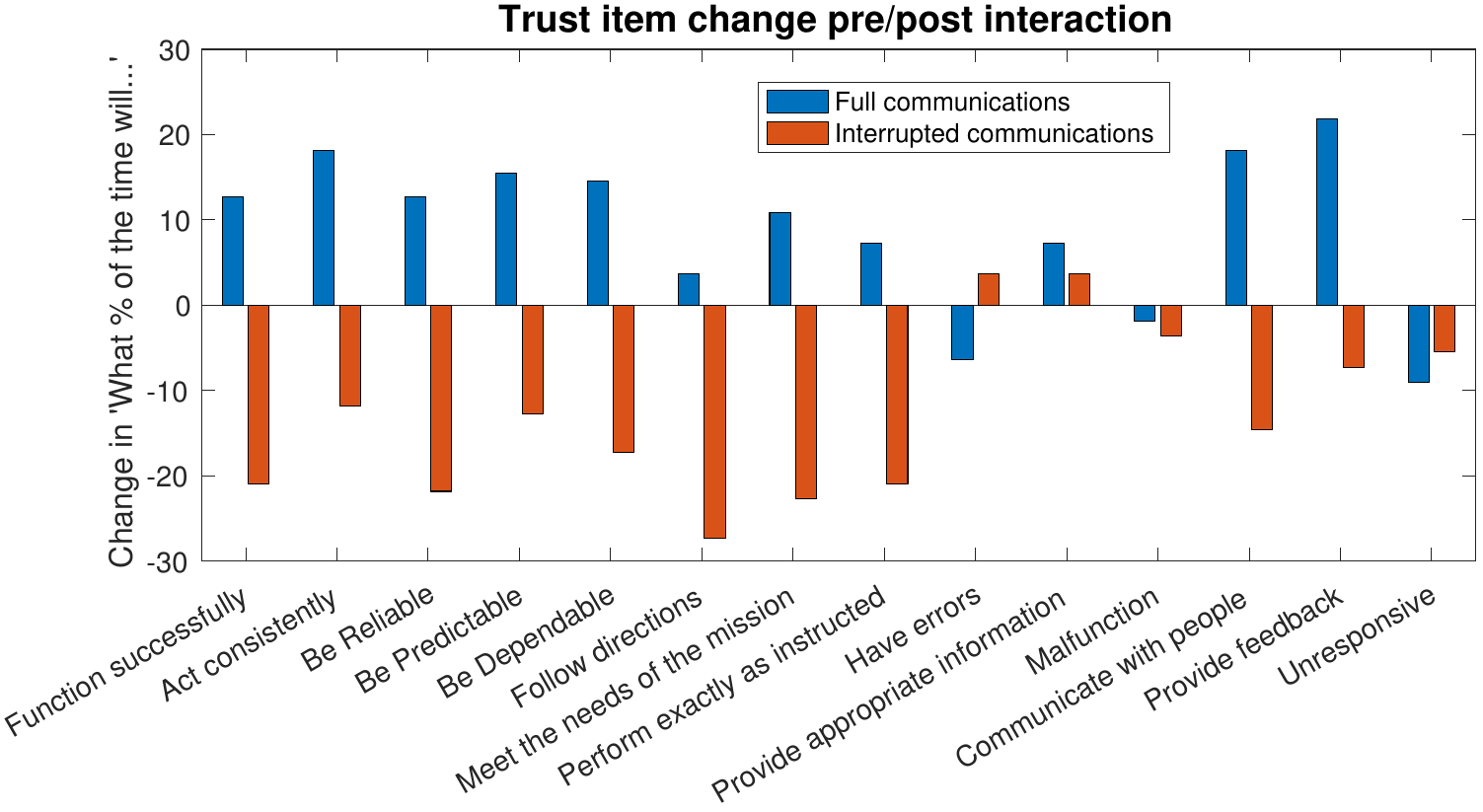}
      \caption{Individual trust survey item changes pre-post interaction}
      \label{fig:trust_item_changes}
   \end{figure}

         \begin{figure}[h]
      \centering
      \includegraphics[width=0.8\columnwidth]{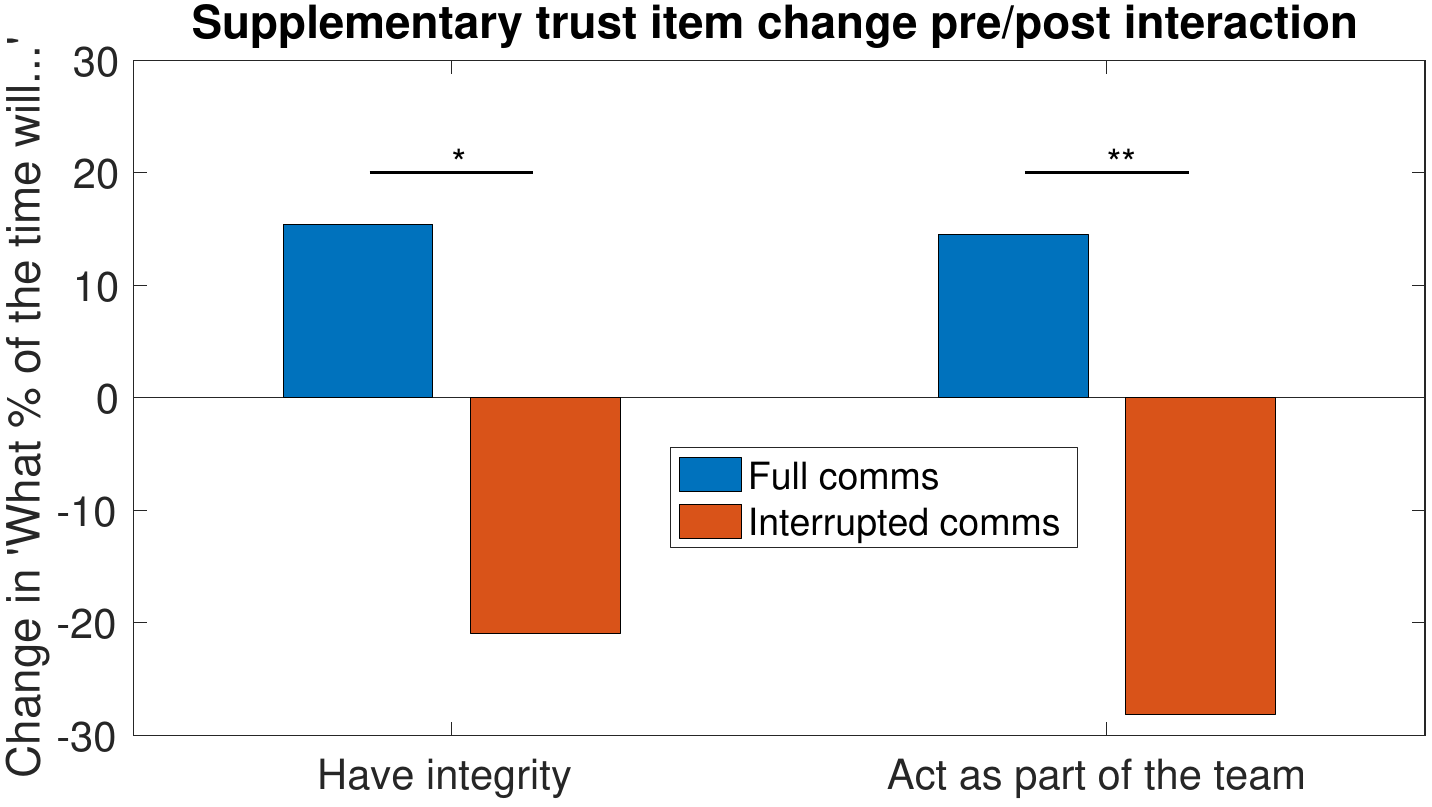}
      \caption{Supplementary trust questions, changes pre-post interaction}
      \label{fig:sup_changes}
   \end{figure}
   
Changes in the individual survey items are shown in Fig.~\ref{fig:trust_item_changes}. For the supplementary questions, `have integrity' and `act as part of the team', the average change pre-post experiment was significantly lower in the `Interrupted Communications' condition (Wilcoxon rank sum tests, $z=2.50, p<0.05; z=3.13, p<0.01$, respectively; Fig.~\ref{fig:sup_changes}).

\subsection{Analysis of Communication and Performance Logs}

We carried out a Two-way Mixed Analysis of Variance (\textit{Condition} as Between Subject and \textit{Robots} as Within Subject factors) for Communications (messages sent) and Tasks Completed with Robot. There was a significant main effect of Condition on Communications [$F(1,21) = 89.96, p<0.001$, observed power = 1] but no effect of Robot.  This suggests that more messages were sent in the `Interrupted communications' condition (18.1 messages to Blue robot, 17.3 messages to Red robot) than in the Full communications condition (7.3 messages to the Blue robot, 5.1 messages to the Red robot).  

There was a significant main effect of Robot on Tasks Completed with Robot [$F(1,21) = 6.124, p<0.05$, observed power 0.656] but no effect of Condition and no interaction effects.  This suggests that participants tended to complete more tasks (shared QR code scans) with the Blue robot (2.7 in the `Full communications' condition, 3.2 in the `Interrupted communications' condition) than the Red robot (2.2 in the `Full communications' condition, 1.2 in the `Interrupted communications' condition).

In terms of Individual tasks, there was a significant difference between Conditions ($t(21) = 2.278, p<0.05$), with participants completing more scans individually in the `Interrupted communications' condition (6.4) than the `Full communications' condition (5.3).  Participants also spent significantly more time completing the `Interrupted communications' condition ($t(21) = 2.537, p<0.05$) than the `Full communications'  condition (8 minutes 28 seconds for the `Interrupted communications' condition versus 6 minutes 15 seconds for the `Full communications' condition).

\subsection{Participant movement behavior}
\begin{figure}
      \centering
      \includegraphics[width=0.8\columnwidth]{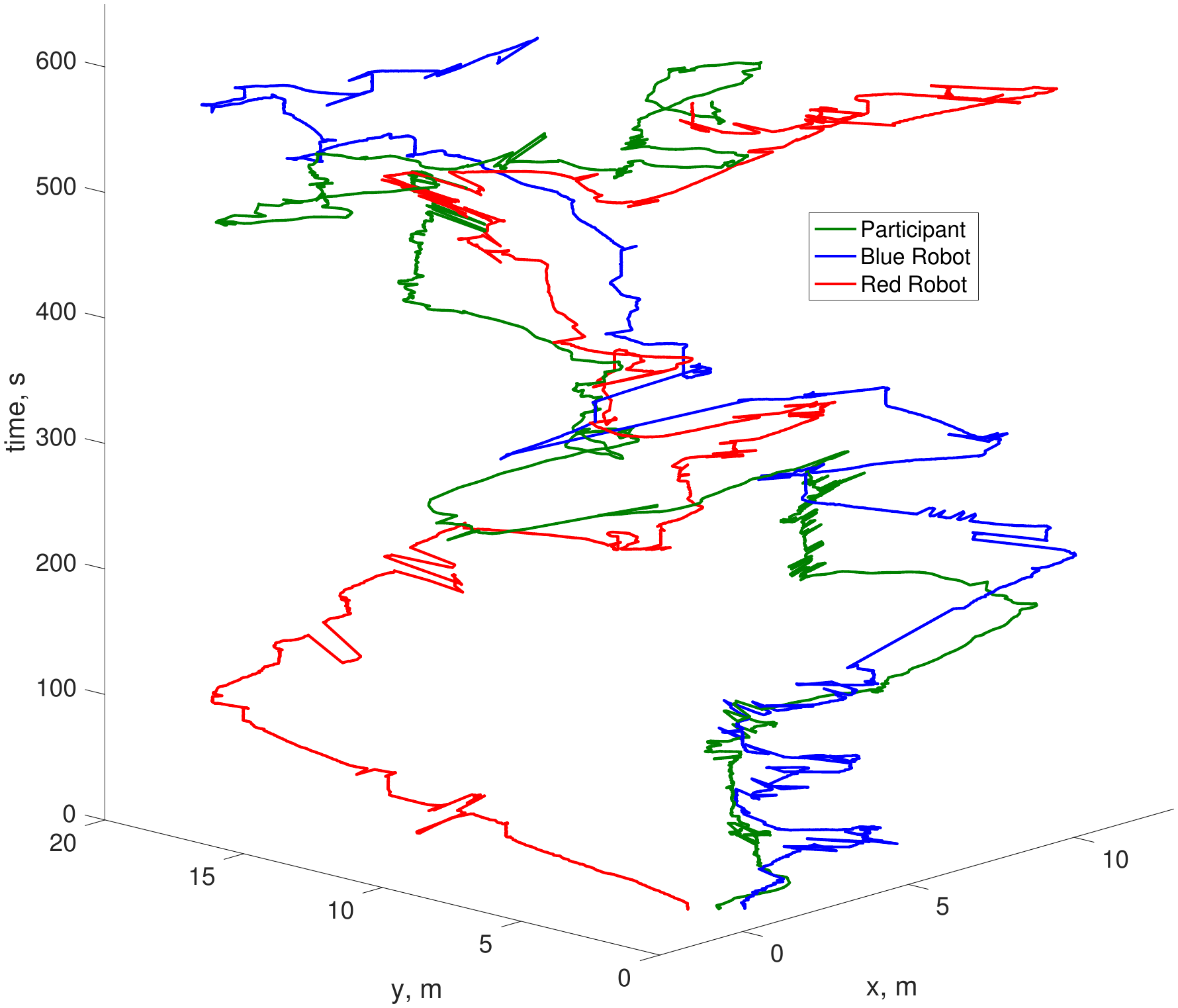}
      \caption{An example of the indoor positioning system tracking trajectories. The z-axis shows time (this trial lasted 642 s or 10 m 42 s). Most of the participants (green line) would search around the space in a counter-clockwise direction, alongside the blue robot initially.}
      \label{fig:trajectory}
   \end{figure}

\begin{figure}
      \centering
      \includegraphics[width=0.8\columnwidth]{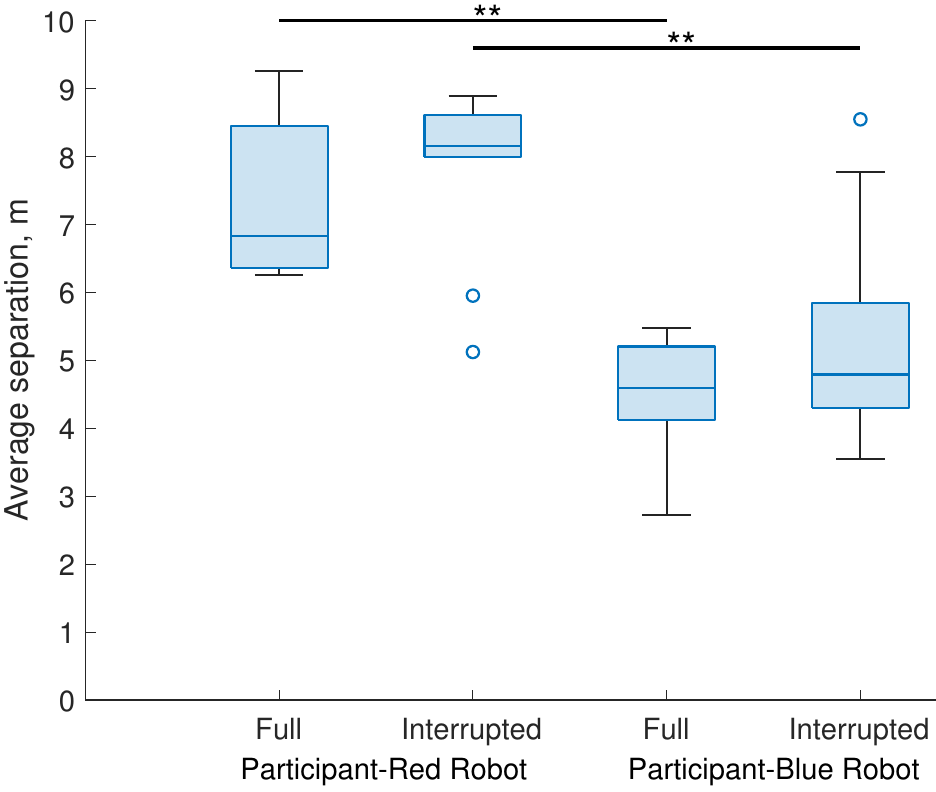}
      \caption{Participants were significantly closer to the blue robot during the trials, in both conditions.}
      \label{fig:separation}
   \end{figure}

Participants were significantly closer to the blue robot in both conditions (Fig.~\ref{fig:separation}, $H(3, n=40)=22.75, p<0.001$, Tukey's HSD probabilities both $p<0.01$). This is apparently because they tended to begin their search in an anti-clockwise direction (e.g. Fig.~\ref{fig:trajectory}). 

\subsection{Interview findings}
Overall, using qualitative content analysis we identified the following themes:

\begin{itemize}
    \item Pre-experiment: expectations; elements of trust and questioning integrity in robots; level of trust
    \item Experiment: perception of robots, automation, robots preference and performance; human-robot teaming; breaking or violation of trust/trust change; trust repair; speed of robots; communication; anthropomorphism.
\end{itemize}

We examine some of these themes here in more detail. 
First we note a finding on experimental realism -- that, while a few participants who had more experience with robots commented that the robots were `simple' and even like a `toy', crucially none shared any suspicion that the `autonomy' was in fact the work of teleoperators. That is to say, the `Wizard of Oz' approach to simulating autonomy was apparently effective in creating a realistic human-robot teaming experience for participants. 

\subsubsection{Elements of Trust}
Out of three suggested elements of trust --- capability, predictability and integrity --- most participants selected capability as the most important one. Here and in subsequent subsections we present some direct quotes to illustrate the richness of the qualitative data. Some suggested capability ``because I still require the machine to finish some tasks I’m not willing to do, or I’m not capable of doing. So if a machine is not good at these matters, then it can digress at this point’'. Indeed, as another participant pointed out, `‘that is the great thing about robots, is they can do things with ease that people can’t, you know...as with any team that you’re in, you want to know that the people you’re working with are going to be able to do that. Everything else kind of comes afterwards’'.

The second biggest group selected predictability --- including reliability, as `‘you don’t know what’s going on in their heads. But as long as they’re doing the same thing every single time, then you know that they’re doing it right, as long as they do it right… every time’'. This is especially important in situations like the one in the experiment, as `‘them actually kind of going in and like to kind of dangerous areas is a good idea. But like, if it’s, it’s whether or not they’re predictable, if that makes sense`‘.

Integrity was, as expected, the most contested issue when it comes to trust elements. While many questioned the use of integrity in HRT, around a quarter of participants singled out integrity as the most important element of trust in HRT. Among the participants who singled out integrity, the requirement for integrity was underpinned by the inherited fear of robots and the potential misalignment of robot goals with human preferences and goals. In addition to fear and potential mistrust in robots’ benevolence, integrity was singled out as the most important element of trust because of the lack of communication with robots. As one participant noted, the lack of direct verbal communication with robots created a greater sense of doubt in the robots' ability to perform as expected: ``you can’t really communicate with robots... So I guess it is that level of like expecting there to be a slip up somewhere of like not, not quite being in sync’'.

\subsubsection{Swift Trust}
When we look into the constructs of swift trust, both imported information and surface level cues were clearly present as influential factors in the interviews, both positively and negatively affecting initial trust. Several participants voiced opinions on the robots' appearance and their ideas of what a robot for such a task should be. One person was dismissive of the robots based on their appearance, `‘[t]hey are just some toys, actually. So, because you’re just typing very simple instruction to let it to move forward like this and it only have two directions to move, either forward or turn back, so I don’t think that is a robot, technically speaking`‘. 

One participant explained that they had a good expectations of the robots' capabilities based on the university behind the study: `‘I wouldn’t have thought you’d have brought anybody in to do this what I did if you thought it might fail. You must have had faith in that machine… Not unless you’ve done it first, yeah`‘. Another participant referred to confidence in the engineers behind the robots: ``I also like naturally trust in people that have built these kind of things that I imagine, you know, a lot of thought and intelligence has gone behind them... I looked at them as well and I had them up and they, they look like little Mars Rovers, like they looked like very well made and also because I was shown the example of how it worked.''

The robots’ performance was often linked to their speed. Most participants commented they were ‘quite slow’. Speed was at times a decisive factor in choosing which robot to call. As one participant observed, ``Yeah, I think there were a couple of times where the blue one was responsive, but it was almost like half the speed of the red one so I was like, looking at the red one like, oooh, should have gone with the red one’'. Speed, especially combined with how the perceived task should be completed, often led participants to single out a favorite robot: ``Yeah, red was one that was better than the other. So, the blue one was less, maybe it still reliable, but it was a bit slow and then at points it seemed like it was moving, not, it was moving around the space in a way that wasn’t completing the mission.'' It is interesting to note that such significant differences in were sometimes perceived by participants, because while there may have been slight variations the actual speed was substantially similar (identical hardware). This may be the result of the experiment design to have limited explicit communications (even in the `Full Communications' condition) and thus an extra vigilance by participants in observing the robots' behavior. 

\subsubsection{Choice of Robot}
Regarding robot preference --- which robot did they choose to call when needing a joint H--R read of the QR code --- most participants indicated the proximity of the robot as the critical factor in decision-making. The perception of the red robot was that the robot had just ``gone off to the left, and I was going the same sort of direction as the blue. So, I initially relied on a blue one'’. Another pointed out that blue was a favorite, because it was ``closest. He followed me the same way around, whereas red went the other way around’'. The proximity of the robot made them a ``part of the team... something that I could use to get the job done’'.

\subsubsection{Interrupted Communications and Trust Damage}

When robots did not respond to humans’ commands, some participants held them to be ‘rude’ as the ‘Robot is unavailable’ message that accompanied the `Interrupted communications' condition was deemed disrespectful:  

``I thought they were doing something selfish... I’m in charge. What I asked them to do is a priority over what they were doing... I mean, I am not a manager, so maybe that’s not correct. I mean, they are autonomous, whereas I’m not. I understand the mission brief, whereas they are kind of just doing their thing. So, I have a priority; it takes precedence over whatever they are doing''.  

Others, however, thought that robots were probably ‘`doing something more important’', based on independent and critical thinking or simply doing their job.

Participants’ reaction to the communications interruption was within the range of ``I just stood there waiting for it and kept calling --– pressing the button’', `‘I had to wait for the red robot to come around, which made me think actually, I’ve got a [blue] robot right next to me... why can’t it come and help me?’' to ‘`I just carry on my route’' and `‘go back to scan the [joint ones] later’'. The most common first response was to try calling the other robot.

Some participants talked about the break of trust in real-life terms (what would this mean in the real-life situation):  ``[The robot] didn’t come to help us. No. If that was a real gas leak, I’d be dead. Yeah... I pressed the button twice, I think three times... Two times too many if you were in the fire... Once, once is OK, twice is like, uh-huh. Once you’ve broken the trust, there is no trust''. This is consistent with past findings of decreased trust in robots if they perform poorly in time-critical, high-risk situations \cite{Robinette2017}. Another said ``I would say if you were to sort of like use it as your car. If I get into my car, then it takes 15 minutes for me to start, that's not a reliable car... I would say they failed, OK... That’s not good enough.''

Questioning robots’ integrity occurred after several repeated failures in communication: ``It started off being about functionality. I was like, oh, there's there's some kind of communications error. Yeah. And therefore, you know, there's some kind of signal problem or it's currently occupied in a task and it hasn't been programmed to answer two things at once. So that was my initial thought. But when that started happening three or four times... it suddenly became less about capability and I started to wonder whether there was an integrity issue.''

\subsubsection{Trust Repair}

Trust repair was an important theme. Here, the nuances in respondents’ answers varied, although a majority suggested that repairing trust is possible with proper (verbal or non-verbal) communication or explanation, even when there are multiple moments of breaking the trust.

Some participants suggested that post-break of trust when robots started ‘working again’ didn’t repair the trust in robots, although one participant did not use the term ‘trust’ but ‘relying’ on a robot. The argument was that ‘trust’ is more of a human quality ``so I don’t know if I give trust, but I do rely on things, I rely on technology a lot to work, to let me do my job.''

Others, on the other hand, felt that after the communications interruption, if robots responded to their call, they could trust them again, especially if the communication channels were working. The level of trust, however, was not the same as at the beginning of the experiment: 

``It went up [after the robot started] working. It went up. I mean on the one hand I mean the system had demonstrated integrity by giving me the right messaging. So that was an improvement. Should have happened before. So, it didn't go up entirely, maybe not as high as it would have done, but, yeah.'' Another said, ``I would probably … I still wouldn't go back to full trust, but I'd go back to like half trust because I know that [the robot] has the ability to suddenly cut out for a large period of time.'' Overall, the interviews indicated that in the absence of clarity in communications, users lose trust quickly when robots do not work correctly.

\section{Discussion \& Future Work}
From the analysis of the trust measures, participants in the `Interrupted communications' condition were significantly more likely to report a decrease in their trust compared with participants in the `Full communications' condition.  This indicates the intervention provoked the intended effect, i.e., interrupted communications led to a reduction in reported trust. From the analysis of the communications and performance logs, interrupted communications had an impact on participants through an increase on messages sent to the robots, longer time to complete the task, and more individual QR codes scanned.  Scanning of shared codes did not seem to be affected by Condition but was affected by Robot (with participants undertaking more shared QR code scans with the Blue robot). One explanation is that, in this experiment, participants were generally committed enough to the task to return to completing the scanning of shared codes once the `Communications restored' message had come through at $t=6$ minutes. Initial analysis of the tracking data shows that the path of the blue robot tended to follow that of the human, and the path of the red robot tended to differ from that of the human. This implies that trust might be mediated by proximity, because the Blue robot (moving in line with the human) tended to be the one that human sought to work with more than the Red robot (operating further from the human). 

Given the relative lack of difference in messages to the two robots, proximity of the robots seemed to have less effect on messages than the interruption to communications. These quantitative metrics indicate some promise of being used to measure and manage human-robot trust online during future ad hoc teaming missions. Interestingly, in this multi-robot HRT, participants developed some sense of robots as individuals they preferred to work with (e.g. the blue robot) and also as a single `system', and we intend to examine this distinction as team sizes scale in future work \cite{Walliser2023}.

From initial analysis of the post-experiment interviews, we can say that participants were aware of the robots' behavior in the interrupted communications condition, although they offered different explanations of this, e.g., from rudeness or selfishness of the robot's part, to performing other, more critical tasks, to technology failure (``some kind of signal problem''). In terms of our trust description, capability and predictability were agreed to be important contributors (which accords with the bulk of literature of this topic, e.g., see \cite{Hancock2020} for a review). There was some debate on the meaning of integrity, although as noted above, some participants believed that the robot  was behaving in a `selfish' manner in the `Interrupted communications' condition (the rating of `have integrity' and `act as part of the team' changed negatively in this condition). Communication (or lack thereof) was highlighted as central by those who raised integrity as a key element of trust, and as critical for effective trust repair. Explainability has previously been highlighted as key for effective human-AI ad hoc teaming \cite{Paleja2021}.

The post-experiment interviews with participants indicated the relevance of both surface-level cues and imported information in forming initial trust in the system \cite{Haring2021,Patel2022}. Participants mentioned factors such robot appearance and movement speed, and imported information such as confidence in the university institution running the experiment. This points to the influence of the individual or organization deploying the robot, termed the `Deployer' in the Social Triad Model proposed by Cameron et al. \cite{Cameron2023}. In designing robot systems with swift trust in mind, given the reliance on surface level cues such as robot appearance, it is also important to be aware of how different robot forms can influence expectations of capability (e.g. \cite{Hancock2011,Haring2018}.

In future studies, we will be exploring the question of integrity and ethical decision-making in HRT as well as communication in trust repair.  A key theme in the post-experiment interviews centered around the repair of trust and how this related to dependency and reliance on technology. In the context of this experiment, it appears that the interrupted communications and the resumption of response from the robots was enough for some participants to allow `trust' (or, for some, `reliance') to be (partly) regained. There was noteworthy variation across participants for what was deemed a sufficient level of communication or explanation of the problem during and after the interruption. Trust restoration might be the result of the instrumental nature of the tasks and future studies will consider other ways in which trust can be impaired; for example, by increasing the negative impact of robot activity on the participant or by making the robot's behavior appear more selfish to the participant.

\begin{acks}
This work was funded by the Engineering and Physical Sciences Research Council (grant EP/X028569/1, ``Satisficing Trust in Human-Robot Teams''). We thank Sparks Bristol, UK, for hosting the experiment.
\end{acks}

\bibliographystyle{ACM-Reference-Format}
\bibliography{tas24-19}

\end{document}